\pdfoutput=1

\documentclass[11pt]{article}

\usepackage[]{acl}

\usepackage{times}
\usepackage{latexsym}
\usepackage{booktabs}
\usepackage{amsmath}
\usepackage{multirow}
\usepackage{hyperref}
\usepackage[normalem]{ulem}
\useunder{\uline}{\ul}{}
\usepackage[pdftex]{graphicx}
\usepackage{tabularx}
\usepackage{xspace}
\usepackage{algorithmicx}
\usepackage{algorithm}
\usepackage{algpseudocode}
\usepackage{dsfont}
\usepackage{amssymb}

\newcolumntype{L}{>{\scriptsize}l}

\usepackage[T1]{fontenc}

\usepackage[utf8]{inputenc}

\usepackage{microtype}

\usepackage{inconsolata}

\newcommand\sadiri{\textsc{Sadiri}\xspace}
\newcommand\iraa{\textsc{Sadiri-v2}\xspace}
\newcommand\reddit{Reddit\xspace}
\newcommand\realnews{RealNews\xspace}
\newcommand\bookcorpus{BookCorpus\xspace}
\newcommand\goodreads{GoodReads\xspace}
\newcommand\amazon{Amazon\xspace}
\newcommand\gmane{Gmane\xspace}
\newcommand\wikiDiscussions{WikiDisc}
\newcommand\pubmed{PubMed\xspace}
\newcolumntype{Y}{>{\raggedright\ttfamily\arraybackslash}X}

\title{Cross-Genre Authorship Attribution via LLM-Based Retrieve-and-Rerank}

\author{Shantanu Agarwal, Joel Barry, Steven Fincke, Scott Miller\\
        Information Sciences Institute \\ University of Southern California \\ \texttt{\{shantanu,joelb,sfincke,smiller\}@isi.edu}}

\begin{document}
\maketitle

\begin{abstract}
Authorship attribution (AA) is the task of identifying the most likely author of a query document from a predefined set of candidate authors. 
We introduce a two-stage retrieve-and-rerank framework that finetunes LLMs for cross-genre AA.
Unlike the field of information retrieval (IR), where retrieve-and-rerank is a de facto strategy, cross-genre AA systems must avoid relying on topical cues and instead learn to identify author-specific linguistic patterns that are independent of the text's subject matter (genre/domain/topic).
Consequently, for the reranker, we demonstrate that training strategies commonly used in IR are fundamentally misaligned with cross-genre AA, leading to suboptimal behavior.
To address this, we introduce a targeted data curation strategy that enables the reranker to effectively learn author-discriminative signals.
Using our LLM-based retrieve-and-rerank pipeline, we achieve substantial gains of 22.3 and 34.4 absolute Success@8 points over the previous state-of-the-art on HIATUS’s challenging HRS1 and HRS2 cross-genre AA benchmarks. 

\end{abstract}
\section{Introduction}

The field of authorship analysis has a long history, dating back at least as far as 1440 when Lorenzo Valla used textual analysis to prove that the \textit{Donation of Constantine} was a forgery \cite{Valla1922}.
It has played a central role in determining the authors of significant works such as the \textit{The Federalist Papers} \cite{Mosteller1984AppliedBA}, as well as contributing to the resolution of criminal cases like the \textit{Unabomber} bombing investigation \cite{Leonard2017AALaw}.

Due to its long-standing history and continued research interest, authorship analysis has been explored across a wide range of settings.
These include variations in task formulation (authorship attribution vs. verification) \citet{Tyo2022OnTS}, text length (short \citet{5615152} vs. long \citet{nguyen-etal-2023-improving}), languages \citet{israeli-etal-2025-million}, domain mismatch (e.g., differences in topic or genre between training and test data) \citet{wegmann-etal-2022-author}, author set assumptions (closed-set \citet{10.1007/978-3-030-49161-1_22} vs. open-set \citet{10.1007/978-3-030-86337-1_15} scenarios), size of the test corpus \cite{rivera-soto-etal-2021-learning}, etc.
Given these nuanced distinctions, it is important to clearly define the scope of this work.

We address the task of authorship attribution and target it as a \textit{ranking task}: given a query document with unknown authorship, the objective is to rank candidate documents by their likelihood of sharing the same author.
All documents in the test corpora are in English.
Our approach generalizes across multiple datasets, spanning long documents (median of $\sim600$ words per document) and medium-length documents (median of $\sim200$ words per document).
The setting is cross-genre, where the query and the relevant candidate document differ not only in topic but also in genre - for example, a news article in politics as the query and a StackExchange post about literature as the candidate (see Table \ref{tab:cross_genre_docs}); the system must generalize to authors unseen during training and therefore cannot rely on author-specific classifiers; at least one candidate document is guaranteed to have the same author as the query; and the candidate author pool is large, comprising tens of thousands of authors.

Building a ranking system that scales to a large candidate pool requires both efficiency and accuracy.
To achieve this, the field of IR commonly adopts a two-stage approach \cite{10.1007/978-3-030-72240-1_26, 10.1145/3626772.3657951}: an efficient retrieval stage followed by a more accurate but compute-intensive reranking stage.
In this work, we extend this paradigm to AA, focusing specifically on the cross-genre setting.
The first stage involves retrieval via a bi-encoder where each document is independently encoded into a vector representation, and the likelihood that two documents share the same author is quantified using the dot product of their vectors.
In the second stage, we use a cross-encoder to rerank the retrieved documents.
The reranker takes both the query and a candidate document as input to directly compute a relevance score.
Due to its joint modeling of the query-candidate pair, the reranker achieves higher accuracy than the retriever, but at the cost of greater computational expense.

Our main contributions are as follows:
\begin{itemize}
\item{
Integrating LLMs: Inspired by the effectiveness of LLMs in IR \cite{10.1145/3626772.3657951}, we leverage them for both the retriever and the reranker.
While prior work has primarily focused on leveraging LLMs' zero/few-shot prompting \citep{hung2023wrotewhypromptinglargelanguage, Huang2024CanLL}, to the best of our knowledge, ours is the first to explore fine-tuning LLMs specifically for AA.
}
\item{
Effective reranker for AA: While prior work has framed AA as a ranking task \citep{rivera-soto-etal-2021-learning, fincke2024separatingstylesubstanceenhancing}, existing approaches have relied primarily on a retrieval stage alone.
In this work, we advance this line of research by introducing a reranking component.
We show that building an effective reranker for cross-genre AA is non-trivial and we address this challenge through targeted training data curation.
}
\end{itemize}
 
We name our system \iraa\footnote{The original \sadiri\cite{fincke2024separatingstylesubstanceenhancing} was a RoBERTa \cite{liu2019robertarobustlyoptimizedbert} based retriever only model and is one of our baselines against which we compare \iraa.}.

\section{Methodology}

We formalize the AA task in Section \ref{sec:method:task_def}. 
We describe our retriever and the reranker in Sections \ref{sec:method:retriever} and \ref{sec:method:reranker} respectively.

\subsection{Task definition}\label{sec:method:task_def}
The objective of an AA system is to give a likelihood score for two documents to have the same author \cite{rivera-soto-etal-2021-learning}. 
Concretely, we have a set of query documents $\mathcal{Q} = \{q_{1}, q_{2}, \cdots, q_{|Q|}\}$, where $q_{i}$ is the $i^{th}$ query document and $|Q|$ is the total number of documents in the query set.
Similar to the query set, we also have a set of candidate documents, $\mathcal{C}$. 
For each document entry in the query collection, the AA system needs to rank the candidate set such that candidate documents which are likely to have the same author as the query document are ranked higher \cite{rivera-soto-etal-2021-learning}. 
For each query document, $q_{i}$, we assume that there is at least one candidate document, $c_{j}$, called the needle, which has the same author as the query.
Documents in the candidate set that are not written by the author of $q_{i}$ act as haystack (or distractors).

We focus on the setting where the query and needle documents are guaranteed to differ both in genre and topic.
This challenge is compounded by the presence of haystack documents that are semantically similar to both the query and the needle (see Table \ref{tab:cross_genre_docs}).
To succeed, an effective AA system must ignore topical cues which could lead to matching with unrelated but topically similar documents while accurately capturing authorial features that link the query to the needle \cite{wegmann-etal-2022-author}.

\subsection{Retriever}\label{sec:method:retriever}
\textit{Architecture}: 
The first stage of our two-stage pipeline prioritizes efficiency and uses a standard bi-encoder architecture for retrieval.
Unlike in prior works on AA, we fine-tune LLMs to serve as the retriever.
Given a document $d$, the transformer LLM outputs a sequence of hidden states $(h_d^1, h_d^2, \ldots, h_d^L)$, where $h_d^i$ is the final-layer hidden representation of the $i^\text{th}$ token, and $L$ is the tokenized document length. We apply mean pooling over these token representations to obtain a fixed-length vector: $v_{\text{agg}}(d) = \text{mean}(h_d^1, h_d^2, \ldots, h_d^L).$
This vector lies in $\mathbb{R}^E$, where $E$ corresponds to the hidden dimension of the transformer. A learnable linear projection is then applied to produce the final document embedding:
\begin{equation}
v(d) = W v_{\text{agg}}(d) + b,
\label{eq:retriever_score}
\end{equation}
where $W \in \mathbb{R}^{D \times E}$ is a projection matrix and $b \in \mathbb{R}^D$ is a bias vector. In our experiments, we set $D = E / 2$.

For training, we construct each batch with $N$ distinct authors, including exactly two documents per author, resulting in a total of $2N$ documents per batch. The bi-encoder is trained using supervised contrastive loss \cite{10.5555/3495724.3497291}, $l = \frac{1}{2N}\sum_{q=1}^{2N}{l_q}$ where:
\begin{align}
\label{eq.contrastive}
    l_{q} &= -\log\frac{\exp{\left(s(d_{q}, d_{q}^{+})/\tau\right)}}{\sum_{d_{c}\in \{d_{q}^{+}\}\cup D^{-}}\exp{\left(s(d_q, d_c) /\tau\right)}}
\end{align}
In the above equation, $s(d_q, d_c)$ denotes the score for a query and a candidate document to share the same author, $d_{q}^{+}$ is the (positive) document in the batch that is written by the same author as $d_q$, $D^{-}$ denotes the set of negative documents, i.e. documents not authored by the author of $d_q$, and $\tau$ is a positive scalar (temperature) hyperparameter.
For the bi-encoder, we calculate the score by taking the dot product between the two document vectors, i.e. $s(d_q, d_c) = v(d_q) \cdot v(d_c)$.
 
\textit{Training batches with hard negatives}: 
For a document $d_q$, a negative document $d^{-}\in D^{-}$ is considered hard if $v(d_q) \cdot v(d^{-})$ is large.
Hard negatives lead to a larger denominator in contrastive loss and thus contribute to faster model convergence.
The importance of hard negative documents is extensively studied in the representation learning literature such as in IR \cite{karpukhin-etal-2020-dense}.
We adopt \sadiri’s \cite{fincke2024separatingstylesubstanceenhancing} strategy for constructing training batches using in-batch negative sampling, where $D^{-}$ includes all $2N - 2$ documents in the batch which are not authored by the writer of $d_q$.
The training documents are first clustered using $k$-means based on the cosine distance between their vectors, and each cluster is assigned to a batch.
The underlying intuition is that documents grouped together by clustering, when not sharing the same author, tend to be hard negatives for each other.
See Appendix \ref{appendix:lora} for details on how the batches are constructed via clustering.

\subsection{Reranker}\label{sec:method:reranker}
\textit{Architecture}: 
We use cross-encoders as our reranker.
Given a query and a candidate document, the model jointly encodes the pair to produce a similarity score reflecting the likelihood of shared authorship.
This joint encoding enables the reranker to have higher discriminative power than the retriever.

Because the reranker jointly encodes the query and the candidate documents, its run-time complexity is $|\mathcal{Q}| \times |\mathcal{C}|$, where $|\mathcal{Q}|$ and $|\mathcal{C}|$ are the number of documents in the query and the candidate collections respectively \footnote{On the other hand, the first-stage retriever's run-time complexity is linear in the number of documents in the collection: $|\mathcal{Q}| + |\mathcal{C}|$.}.
Because of the reranker's quadratic run-time complexity, it is infeasible to rank all candidate documents for each query input, and thus, the reranker is applied only to the top-$k$ candidates retrieved in the first stage.

We fine-tune LLMs as cross-encoders by concatenating the query and candidate documents using a special delimiter.
The reranker computes an authorship similarity score as:
\begin{equation}
\label{eq.reranker_score}
s(d_q, d_c) = W h^{L}_{(d_q, d_c)},
\end{equation}
where $h^{L}_{(d_q, d_c)} \in \mathbb{R}^{E}$ is the final-layer hidden state of the last input token, $E$ is the LLM's hidden dimension, and $W \in \mathbb{R}^{1 \times E}$ is a learned projection matrix.
The reranker is trained using the contrastive loss in Eq. \ref{eq.contrastive}, with similarity scores computed as per Eq. \ref{eq.reranker_score}.

\begin{figure}[t!]
    \centering
    \includegraphics[width=0.97\linewidth]{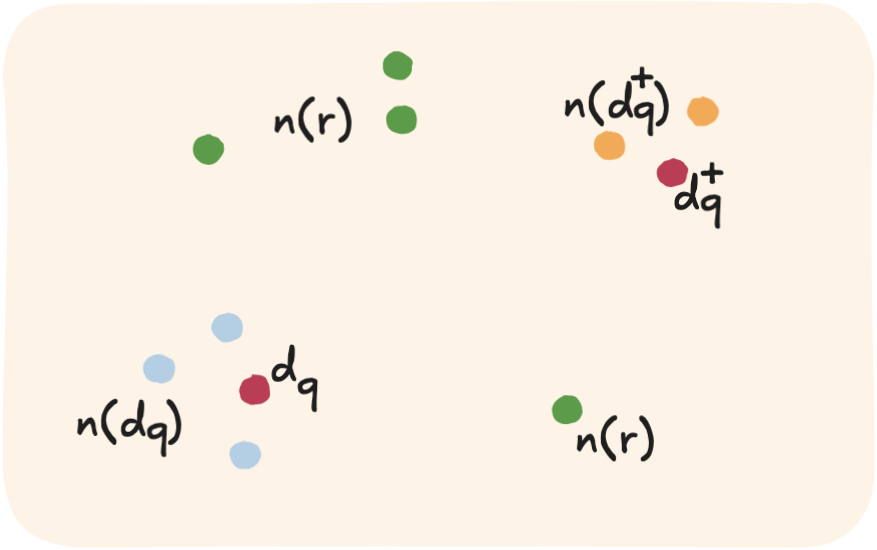}
    \caption{
    Negative documents, i.e. documents not written by the author of the query document, can be categorized into three distinct groups:
    $n(d_q)$ are documents close to the query document, $n(d_q^+)$ are close to the positive document and $n(r)$ are random documents with no particular bearing to either $d_q$ or $d_q^+$.
    }
    \label{fig:reranker_hn}
\end{figure}

\textit{Hard negatives to train the reranker}:
We construct each training batch using $N$ authors, with two documents per author: one serving as the query document ($d_q$) and the other as the positive candidate ($d_q^+$).
For each query, $m$ negative documents not authored by the query author are selected to form $D^-$.
Thus, in total, each $d_q$ is compared against $(1 + m)$ candidate documents.

Based on its similarity to the query and the positive document, each negative document falls into one of three categories:
$n(d_q)$ - close to the query document, $n(d^+)$ - close to the positive document, or $n(r)$ - random. 

Note that this categorization of negatives is fundamentally different from the choice of negatives available in IR.
In IR, the negatives for the reranker are typically drawn from the retriever’s top-ranked candidates \cite{10.1007/978-3-030-72240-1_26}, i.e. those with a high $s(d_q, d_c)$ score in Eq. \ref{eq.contrastive}.
In addition, IR retrievers typically do not explicitly model the closeness of candidate documents to the positive, i.e. $s(d_q^+, d_c)$ is not available.
This distinction arises from a fundamental asymmetry in IR: queries are often short and structurally different from candidate documents.
In contrast, AA involves comparing two documents, making the query and the candidate symmetric in nature.

We demonstrate that the reranker’s performance is strongly influenced by the distribution of negative samples used for training. 
Sampling exclusively from $n(d_q)$ (analogous to IR-style retrieval) yields insufficient contrastive signal for learning robust cross-genre authorship representations.
We also demonstrate that sampling negatives from all three categories, $n(d_q) + n(d_q^+) + n(r)$, leads to the best cross-genre system performance.

\section{Experiments}
In this section, we evaluate our retrieve-and-rerank approach on the cross-genre authorship attribution task. 
Section \ref{subsection:train_dataset} describes our training data selection strategy for effective cross-genre AA,
section \ref{subsection:test_dataset} describes HIATUS's cross-genre test corpus, section \ref{subsection:metrics} outlines the evaluation metrics, section \ref{subsection:baselines} presents the baseline methods, and section \ref{subsection:implementation_details} provides implementation details.

\subsection{Curating training data for cross-genre AA}\label{subsection:train_dataset}
Building a strong cross-genre AA model requires guiding it to rely on authorial style rather than content.
This is typically achieved through two strategies: (1) carefully selecting training data to discourage content-based matching \cite{wegmann-etal-2022-author, fincke2024separatingstylesubstanceenhancing}, and (2) introducing inductive bias by masking content-bearing words, such as replacing them with their corresponding POS tags \cite{10.1162/tacl_a_00610}.

We adopt the first strategy.
Starting with web-scraped documents across 8 diverse genres, we apply the following filtering steps: remove documents with fewer than 350 words\footnote{Words are defined as white-space delimited spans.}; discard authors with fewer than 2 or more than 50 documents\footnote{The upper bound helps eliminate bot-generated content.}.
For each author, we calculate the SBERT-based\footnote{\href{https://huggingface.co/sentence-transformers/all-mpnet-base-v2}{https://huggingface.co/sentence-transformers/all-mpnet-base-v2}} \cite{reimers-2020-multilingual-sentence-bert} cosine similarity between all document pairs and add the most dissimilar document pair to our training collection if the cosine similarity is less than a predefined threshold. 
To ensure a fair comparison, we select \sadiri’s threshold of 0.2.
SBERT based filtering ensures that $d_q$ and $d_q^+$ are topically dissimilar; hence the bi-encoder/reranker models must learn features other than topic when distinguishing the positive from the negative documents.
The final training set contains 395,118 documents from 197,559 authors, with exactly two documents per author. See Appendix \ref{appendix:training_data} for additional details.

\subsection{Test Dataset}\label{subsection:test_dataset}
HIATUS\footnote{https://www.iarpa.gov/research-programs/hiatus} is an IARPA\footnote{https://www.iarpa.gov/} initiative focused on developing human-usable systems for attribution of authorship and protection of author privacy. 
We test the robustness of our approach on HIATUS’s HRS1 and HRS2 English test sets\footnote{All test data and scripts to create splits used in this work are publicly available through https://www.iarpa.gov/research-programs/hiatus}, which contain long and medium-length documents, respectively. 
HRS1 includes only documents with at least 350 words, with a median length of 581 words. In contrast, HRS2 consists of medium-length documents, each with at least 80 words and a median length of 206 words.
Both HRS1 and HRS2 include query and candidate documents that span five distinct genres.
We provide the distribution of each genre in these datasets in Appendix \ref{appendix:test_data}.
For each genre, two types of document are provided: background documents, scraped from the Web, and foreground documents, written specifically for the HIATUS program.
The foreground authors contribute documents in multiple genres, enabling a controlled and diverse evaluation setting for cross-genre authorship attribution.
The foreground set addresses the difficulty of obtaining cross-genre samples from the same author in naturally occurring Web data.
To highlight the cross-genre variation, we present sample texts from a foreground author in two distinct genres in Table \ref{tab:cross_genre_docs}.

In HIATUS's HRS1 and HRS2 datasets, the queries are always drawn from the foreground collection. 
For each query, the corresponding needle document(s) are guaranteed to be sampled from genres \textit{different} from from that of the query. 
This design enforces a strict evaluation of cross-genre performance.
The needles in addition to the background documents comprise the candidate pool.
Details about the test collection is provided in Appendix \ref{appendix:test_data}.

\subsection{Metrics}\label{subsection:metrics}
We assess system performance with established ranking metrics commonly used in IR.
For all experiments, we report Success@$k$ at $k \in {8, 100}$\footnote{Success@8 is the official evaluation metric for the HIATUS task.}, as well as Mean Reciprocal Rank at 20 (MRR@20).

Success@$k$ measures the proportion of queries for which at least one needle, i.e. a candidate document written by the same author as the query, is ranked among the top $k$ results.
The mean reciprocal rank at $k$ (MRR@$k$) computes the average reciprocal rank of the first correct candidate, considering only the top $k$ ranked documents per query.

\begin{table*}[t]
\centering
\renewcommand{\arraystretch}{1.2}
\begin{tabularx}{\textwidth}{lllY}
\hline
\textbf{Author} & \textbf{Type} & \textbf{Genre} & \textbf{Text} \\
\hline
A & query & Instructions & How to Make a Curtain Tie-Back\newline Introduction: Curtain Tie-Back\newline Curtain tie-backs are a fantastic ... \\
\hline
A & needle & News & Earthquakes in Turkey and Syria\newline An earthquake of the degree of 7.8 recently ... \\
\hline
B & distractor 1 & Instructions & How to make a homemade rug\newline The homemade rug is time costly but with determination, you are ... \\
\hline
C & distractor 2 & News & Media Reportage on Conflict (Russia \& Ukraine)\newline Conflict simply means ... \\
\hline
\end{tabularx}
\caption{Example documents (first few sentences) from HIATUS’s HRS1 collection illustrating the challenge of cross-genre authorship attribution. 
The query and needle come from different genres and topics, requiring the system to capture authorial style rather than topical overlap. 
At the same time, the system must reject distractors that are topically similar to either the query or the needle.}
\label{tab:cross_genre_docs}
\end{table*}

\subsection{Baseline Models}\label{subsection:baselines}
We consider two broad categories of models: IR models and fine-tuned authorship-specific models.

\textbf{IR Baselines}:
IR models provide a reference point for quantifying the improvements gained through task-specific adaptation.
\begin{itemize}
\item{
BM25: We use the BM25 relevance score as a measure of authorship similarity.
Since BM25 relies solely on lexical matching of content-bearing words, we show that it is ill-suited for cross-genre AA, where lexical overlap, especially of content-bearing words, between documents by the same author may be minimal.
}
\item{
SBERT: The authorship similarity score is derived from the cosine between the SBERT \cite{reimers-2020-multilingual-sentence-bert} embeddings of the query and the candidate documents.
Similar to BM25, we show that SBERT underperforms on the cross-genre AA task because the query and relevant candidate documents exhibit substantial topical dissimilarity.
}
\end{itemize}

\textbf{Authorship-specific models}:
In addition to the IR models, we consider \sadiri \cite{fincke2024separatingstylesubstanceenhancing} and LUAR \cite{rivera-soto-etal-2021-learning}; models which are explicitly fine-tuned for the AA task.
\begin{itemize}
\item{
LUAR: LUAR \cite{rivera-soto-etal-2021-learning} is a RoBERTa-base \cite{liu2019robertarobustlyoptimizedbert}  model fine-tuned for AA.
Similar to our approach, LUAR also targets ranking-based evaluation and is trained using the contrastive loss objective (Eq. \ref{eq.contrastive}).
However, LUAR's training curriculum exhibits two key shortcomings: (1) it does not leverage hard negative mining, which is critical for discriminative representation learning, and (2) it is not explicitly optimized for the cross-genre setting.
}
\item{
\sadiri: \sadiri \cite{fincke2024separatingstylesubstanceenhancing} fine-tunes a RoBERTa-large model for AA.
It addresses the two primary limitations of LUAR by incorporating hard negative mining and carefully curating the training data to promote robust cross-genre performance.
Among all baselines considered, \sadiri is the strongest, representing the current state-of-the-art in cross-genre AA.
}
\end{itemize}

\subsection{Implementation Details}\label{subsection:implementation_details}
\textit{Baselines}: For the BM25 experiments, we use the standard hyperparameters: $k_1 = 0.25$ and $b = 0.75$. 
For SBERT, we use the all-mpnet-base-v2\footnote{https://huggingface.co/sentence-transformers/all-mpnet-base-v2} model. 
We use the settings proposed in \cite{fincke2024separatingstylesubstanceenhancing} for \sadiri (Appendix \ref{appendix:lora}).

\textit{\iraa retriever}: Our proposed retriever improves upon \sadiri by replacing the RoBERTa-large bi-encoder with a decoder-only LLM backbone.
We use \sadiri’s default hyperparameters, including hard-negative batching, with the following key efficiency-focused adjustments:
(1) Parameter-efficient tuning: Instead of updating all the LLM weights, we only train LoRA parameters \cite{Hu2021LoRALA}. 
(2) Reduced batch size: We use smaller batches (e.g., 16 authors per batch for 
Qwen\footnote{https://huggingface.co/collections/Qwen/qwen3-67dd247413f0e2e4f653967f} and Mistral-Nemo-Base-2407\footnote{https://huggingface.co/mistralai/Mistral-Nemo-Base-2407} vs. 74 in \sadiri).
(3) Fewer training epochs: We trained for only a single epoch.
(4) Efficient embedding caching: \sadiri’s hard-negative batching necessitates caching the embeddings of the entire training set.
However, with LLMs, this naive approach leads to out-of-memory issues which we overcome by introducing a lightweight caching mechanism optimized for memory efficiency.
For additional details, please refer to Appendix \ref{appendix:lora}

We investigate the effect of model scaling on AA performance using Qwen3 models of varying sizes, with our largest experiment conducted on a 12-billion-parameter Mistral-Nemo-Base-2407 model.

\textit{\iraa reranker}: Like our retriever, the \iraa reranker also finetunes LLMs by using the cross-genre-targeted training collection described in Section \ref{subsection:train_dataset}.
For each author $a$ in the training collection, we randomly label one of the authored documents as the query ($d_q$) and the other as the positive candidate ($d_q^+$).
We then sample $m$ other documents from the training corpus not authored by $a$ to form the negative set $D^-$.
Thus, in total, each $d_q$ is compared against a total of $(1 + m)$ candidate documents.
For all our experiments, we fix the number of negatives to $m = 12$.
An input instance to the LLM consists of a query and a candidate document concatenated using a rarely occurring character as the delimiter\footnote{We use Unicode character U+2980 which never appeared in our training set.}.
Both the query and candidate are tokenized independently with a maximum length of 512 tokens, resulting in a combined context length of 1,024 tokens (excluding special tokens).
For all reranker experiments, we use Mistral-Nemo-Base-2407 as the backbone LLM and optimize only the LoRA parameters during training (See Appendix \ref{appendix:lora}).

As described in Section \ref{sec:method:reranker}, there are three categories from which negative documents can be sampled: $n(d_q)$, $n(d_q^+)$ and $n(r)$.
These categories rely on a measure of closeness to the query ($d_q$) and the positive document ($d_q^+$).
In our experiments, we define closeness using SBERT-based\footnote{\href{https://huggingface.co/sentence-transformers/all-mpnet-base-v2}{https://huggingface.co/sentence-transformers/all-mpnet-base-v2}} cosine similarity.
When negatives are sampled from multiple categories, we sample uniformly across categories, e.g., when sampling from $n(d_q) + n(d_q^+) + n(r)$, we select $m/3$ negatives  from each category.

Training a reranker is computationally more expensive than training the first stage retriever. 
For the retriever, incorporating $n$ authors into a batch requires only $2n$ individual documents as inputs to the transformer. 
By contrast, for $n$ query authors in a batch, the reranker requires $n(1+m)$ query–candidate pairs, which translates to $(1+m)/2$ additional transformer inputs per author compared to the retriever. Furthermore, it must process both the query and the candidate document within the same sequence, effectively doubling the context length relative to the retriever.
Given the significantly higher computational cost, we train the reranker on only 10\% of the query-authors used to train the retriever.

Since the reranker has a quadratic run-time complexity of $|\mathcal{Q}| \times |\mathcal{C}|$, to balance effectiveness with efficiency, in all our experiments, at inference time, the reranker is applied only to the top-$k$ ($k = 100$) candidates retrieved by the first-stage retriever.

\section{Results}\label{sec:results}

\begin{table*}[hbt!]
\small
\centering
\begin{tabular}{llcccccc}
\toprule
\multirow{2}{*}{Framework} & \multirow{2}{*}{Model / Params} 
& \multicolumn{3}{c}{HRS1 / English Long} & \multicolumn{3}{c}{HRS2 / English Medium} \\
\cmidrule(lr){3-5} \cmidrule(lr){6-8}
 & & Success@8 & Success@100 & MRR@20 & Success@8 & Success@100 & MRR@20 \\
\midrule\midrule
BM25   & ---                        & 0.0  & 0.0  & 0.0  & 0.0  & 5.7  & 0.0  \\
SBERT  & all-mpnet-base-v2 / 0.1B & 0.2  & 3.1  & 0.3  & 0.7  & 3.4  & 0.2  \\
LUAR   & RoBERTa-base / 0.1B      & 7.2  & 25.3 & 3.7  & 18.2 & 49.5 & 12.3 \\
\sadiri & RoBERTa-large / 0.3B    & 26.0 & 62.8 & 14.8 & 34.0 & 58.2 & 18.6 \\
\addlinespace
\iraa  & Qwen3 / 0.6B             & 27.5 & 67.3 & 16.8 & 30.6 & 64.2 & 19.4 \\
\iraa  & Qwen3 / 1.7B             & 32.2 & 77.5 & 18.2 & 37.1 & 65.7 & 23.3 \\
\iraa  & Qwen3 / 4B               & 32.7 & 78.7 & 18.7 & 46.2 & 74.7 & 30.6 \\
\iraa  & Qwen3 / 8B               & 35.7 & 83.8 & 21.9 & 51.7 & 77.3 & 33.4 \\
\iraa  & Mistral / 12B            & \textbf{42.1} & \textbf{84.9} & \textbf{24.0} & \textbf{59.9} & \textbf{84.2} & \textbf{45.6} \\
\bottomrule
\end{tabular}
\caption{
\iraa's cross-genre AA performance of retrieval-based models on the HRS1 and HRS2 datasets.
Our results show that scaling model size consistently improves performance, with the 12B Mistral model surpassing previous state-of-the-art, \sadiri, by 16.1 Success@8 points on HRS1 and by 25.9 points on HRS2. 
}
\label{tab:retrieval}
\end{table*}

Table~\ref{tab:retrieval} presents the performance of retrieval models for cross-genre AA. 
BM25 and SBERT are incapable of discriminating authorship style. 
This demonstrates the challenge of AA where models solely relying on content (either lexical or semantic) are unable to effectively perform attribution.
LUAR, which was trained specifically for AA, performs better than these baselines, but still lags behind \sadiri and \iraa due to architectural and training limitations. 
In particular, LUAR relies on RoBERTa-base, uses truncated inputs (32 tokens per document and aggregates tokens across 16 documents for each author), and does not optimize for cross-genre or employ hard-negative batching. 

\sadiri is the current state-of-the-art model for cross-genre AA.
It overcomes LUAR's limitations by targeting cross-genre performance, training efficiently using hard-negative batching and training with maximum context length of 512 tokens using RoBERTa-large.
Building on \sadiri, our \iraa retrievers incorporate LLM backbones. 
The Qwen family of models shows a clear trend: larger models yield consistent improvements, with performance rising from 0.6B to 8B parameters.
The biggest model we build, Mistral, has 12B parameters, yielding an absolute gain of 16.1 Success@8 points for HRS1 and 25.9 points for HRS2.

A key observation across all models and datasets is the consistently large gap between Success@8 and Success@100.
For instance, with the Mistral model, this gap is $42.8$ points $(84.9 - 42.1)$ on the HRS1 dataset.
In other words, while the retriever often fails to place the correct document within the top-$8$, it frequently succeeds in ranking it within the top-$100$.
This motivates our use of a reranker on the top-$100$ candidates, with the goal of narrowing the gap between Success@8 and Success@100. 

Table \ref{tab:reranker} reports \iraa's reranker performance.
For each query document, we reorder the top-$100$ candidates retrieved by our strongest retriever, \iraa/Mistral (the retriever performance is shown in the first row of Table \ref{tab:reranker}).
All rerankers are built on top of a pretrained Mistral model\footnote{https://huggingface.co/mistralai/Mistral-Nemo-Base-2407}.
As discussed in Section \ref{subsection:implementation_details}, when training the reranker, in addition to the single positive document, we compare each query document against $m=12$ negative documents.
The distribution of these $m$ negative documents ranges from $n(r)$, where all negatives are randomly sampled, to $n(r) + n(d_q) + n(d_q^+)$, where equal number ($4$ when $m=12$) of negatives is drawn from each of $n(r)$, $n(d_q)$, and $n(d_q^+)$.

We observe that the strategy used to sample negative training documents has a significant impact on the reranker performance. 
Specifically, we see that the performance of the reranker is best when negative documents are drawn from the distribution $n(r) + n(d_q) + n(d_q^+)$, providing a significant improvement of Success@8 compared to the retriever alone by 6.2 points for HRS1 and 8.5 for HRS2.

\textbf{Sampling only from $n(d_q)$ results in the worst performance}: 
From Table \ref{tab:reranker}, it is clear that selecting negatives only around the query, i.e. selecting from $n(d_q)$, yields the weakest performance across datasets and metrics, even worse than the retriever itself.
The reason is rooted in how training data are constructed: $d_q$ and $d_q^+$ are deliberately chosen to be semantically dissimilar to enforce cross-genre robustness (see Section \ref{subsection:train_dataset}).  
When the reranker is trained to score $d_q^+$ above negatives sampled only from $n(d_q)$, the optimization reduces to learning an anti-topic heuristic: documents semantically close to $d_q$ are always scored low, while those semantically farther away are scored high.  
Thus, this failure mode is an inherent consequence of adapting the reranker to the cross-genre AA setting.

\begin{table*}[hbt!]
\small
\centering
\begin{tabular}{llcccccc}
\toprule
\multirow{2}{*}{Framework} & \multirow{2}{*}{Negative sampling strategy} 
& \multicolumn{3}{c}{HRS1 / English Long} & \multicolumn{3}{c}{HRS2 / English Medium} \\
\cmidrule(lr){3-5} \cmidrule(lr){6-8}
 & & Success@8 & Success@100 & MRR@20 & Success@8 & Success@100 & MRR@20 \\
\midrule\midrule
Retriever  & -                   & 42.1 & 84.9 & 24.0 & 59.9 & 84.2 & 45.6 \\
\midrule
   & $n(r)$                      & 40.2 & -    & 26.9 & 66.4 & -    & 49.1 \\
   & $n(d_q)$                    & \underline{34.8} & -    & \underline{17.7} & \underline{58.8} & -    & \underline{30.6} \\
   & $n(d_q^+)$                  & 44.1 & -    & 27.6 & 66.7 & -    & \textbf{54.2} \\
Reranker   & $n(r) + n(d_q)$     & 47.0 & -    & 30.8 & 66.7 & -    & 51.6 \\
   & $n(r) + n(d_q^+)$           & 44.0 & -    & 27.7 & 67.4 & -    & 50.9 \\
   & $n(d_q) + n(d_q^+)$         & 46.6 & -    & 30.6 & 68.0 & -    & 53.5 \\
   & $n(r) + n(d_q) + n(d_q^+)$  & \textbf{48.3} & - & \textbf{31.1} & \textbf{68.4} & - & 53.1 \\
\bottomrule
\end{tabular}
\caption{
\iraa's cross-genre AA performance of reranker-based models on the HRS1 and HRS2 datasets.
We take the output of our strongest reteiver (Mistral / 12B) and rerank its top-$100$ retrieved candidate documents.
Highest and lowest values under each column are in bold and underlined respectively.
The table shows the variation in reranker performance across different negative document sampling strategies.
Sampling negatives around the query ($n(d_q)$) results in the worst performance, scoring even below the retriever.
$n(r) + n(d_q) + n(d_q^+)$ performs the best and improves over the retriever by 6.2 Success@8 points on HRS1 and 8.5 points on HRS2.
}
\label{tab:reranker}
\end{table*}
\section{Related Work}


\textit{LLMs for authorship analysis}: 
\citep{hung2023wrotewhypromptinglargelanguage, Huang2024CanLL} reported that LLMs are capable of determining whether or not two documents are written by the same author (the task of authorship verification).
However, their methodology has three key limitations: 1) they did not test on cross-genre scenario and so their conclusion is likely to be confounded by LLM's ability to pick that same authors tend to write about similar topics, 2) they tested on datasets such as Enron emails \cite{10.1007/978-3-540-30115-8_22} which are very likely to be already present in most LLM's pretraining dataset, and 3) they solely explored LLM's zero-shot ability via prompting.
We address all three limitations; we target cross-genre AA, test on the HIATUS dataset in which the foreground documents are solely written for the purpose of IARPA's HIATUS program and are less likely\footnote{HIATUS prohibits the use of its data in closed-source models such as GPT.} to have already been included in LLM's pretraining collection, and finally, we finetune LLMs to adapt it for the AA task.

\textit{Retrieve-and-rerank for AA}: 
Previous work has framed authorship attribution as a ranking task \citep{rivera-soto-etal-2021-learning, fincke2024separatingstylesubstanceenhancing}, where the candidate pool may include tens of thousands of documents, similar to large-scale settings in information retrieval (IR).
Prior ranking-based approaches to authorship analysis, however, have relied solely on the retrieval component \citep{rivera-soto-etal-2021-learning, fincke2024separatingstylesubstanceenhancing}.
We extend this line of work by developing an effective reranker to significantly improve the system's end-to-end performance.

\textit{Cross-genre authorship attribution}: 
Previous works have explored methods to ensure authorship verification and analysis models encode linguistic style alone and not content/topic/symmantics \cite{10.1145/3339252.3340508, wegmann-etal-2022-author, fincke2024separatingstylesubstanceenhancing}.
We follow \sadiri's approach by training on pairs of an author’s most semantically dissimilar documents to reduce topic bias.
Additionally, to the best of our knowledge, we are the first to address the challenge of building an effective cross-encoder/reranker for cross-genre AA\footnote{Previous works \citep{wegmann-etal-2022-author, fincke2024separatingstylesubstanceenhancing} developed content-controlled authorship analysis models, but they focused on bi-encoders/retrievers rather than cross-encoders/rerankers.}. 

\textit{Negative document sampling in IR}:  
Training IR models benefits substantially from the inclusion of hard negatives \citep{Xiong2020ApproximateNN, 10.1007/978-3-030-72240-1_26}.  
Typically, these negatives are selected to be close to the query, without explicitly considering their distance from the positive document \citep{Xiong2020ApproximateNN}.  
Some studies \citep{10.1609/aaai.v38i8.28779, meghwani2025hardnegativeminingdomainspecific}, however, incorporate the positive document into the sampling process to reduce false negatives; documents labeled as irrelevant to the query but that are in fact relevant which is a common issue in IR training data.  

\section{Conclusion}
This work extends the retrieve-and-rerank framework, a pipeline that is canonically used in IR, to the field of cross-genre AA. 
Leveraging LLMs for retievers enables us to surpass previous state-of-the-art, \sadiri, by a significant margin (16.1 Success@8 points on HRS1 and 25.9 points on HRS2).
We find that building an effective reranker for cross-genre AA is far from straightforward: when hard negatives are sampled using the IR-style strategy of placing them near the query, performance not only degrades, but the reranker scores below the retriever itself.
Our cross-genre targetted reranker delivers substantial gains (6.2 Success@8 points on HRS1 and 8.5 points on HRS2), further boosting performance beyond our already strong retriever.
Taken together, our two-stage retrieve-and-rerank pipeline improves cross-genre HRS1 and HRS2 scores over the previous state-of-the-art, \sadiri, by 22.3 and 34.4 Success@8 points, respectively.

\section*{Limitations}
Improvements made in this work are based on using larger models than previously used for AA.
Training and inference with these LLMs is computationally expensive.
For instance, the Mistral-12B retriever is approximately $20\times$ slower than RoBERTa-large during training and about $6\times$ slower at inference.
Because the reranker jointly encodes each query-candidate pair, its computational expense is higher than the retrievers.
To compensate for this, we train the reranker on only 10\% of our training collection and rerank only the top-$100$ retrieved documents.

Due to hardware constraints, Mistral-Nemo-Base-2407 is the largest model (in terms of number of parameters) that we trained.
Whether the performance of the attribution system further improve upon using even larger models remains an open question.

Selecting negative documents for the reranker requires one to define a measure of closeness.
In this work, we define closeness based on document embeddings derived from an SBERT model. 
In IR, alternative measures of closeness (such as BM25 scores or retriever-derived embeddings \cite{gao2021rethinktrainingbertrerankers}) are also widely used, and exploring these remains an interesting direction for future work.

\section*{Acknowledgments}
This research is supported in part by the Office of the Director of National Intelligence (ODNI), Intelligence Advanced Research Projects Activity (IARPA), via the HIATUS Program contract \#2022-22072200006. 
The views and conclusions contained herein are those of the authors and should not be interpreted as necessarily representing the official policies, either expressed or implied, of ODNI, IARPA, or the U.S. Government. 
The U.S. Government is authorized to reproduce and distribute reprints for governmental purposes notwithstanding any copyright annotation therein. 

\bibliography{anthology,custom}

\appendix

\section{Training data}
\label{appendix:training_data}

We train all our models on data scraped from 8 web sources.
To encourage our AA models to rely on authorial style and not on topic, we include the two most topically dissimilar documents by each training author and include the document-pair only if their SBERT similarity is less than a threshold.
Across all experiments, similarity was computed as the dot product between document embeddings generated by the all-mpnet-base-v2\footnote{https://huggingface.co/sentence-transformers/all-mpnet-base-v2} SBERT model and threshold of 0.2 was used to be consistent with \sadiri. 
The training data is processed by removing personally identifiable information (PII).
Specifically, identifiers pertaining to person names, telephone numbers, email addresses and ip-addresses are replaced by to PERSON, PHONE\_NUMBER, EMAIL\_ADDRESS and IP\_ADDRESS respectively.
Note that removing PII is not only crucial for preserving privacy but also essential for fair AA evaluation, since such identifiers can provide trivial author-specific cues, allowing models to bypass learning stylistic features.
In Table \ref{table:eng_8g}, we list all the data sources along with the number of documents and authors which we used in training of all the \sadiri and \iraa models. 

\begin{table*}[!h]
\centering
\begin{tabular}{@{}llr@{}}
\toprule
source &  description & \# docs / \# authors     \\ \midrule
\realnews & news stories & 191,020 / 95,510 \\
\bookcorpus & full-length books & 66,786 / 33,393 \\
\reddit & reddit.com entries & 55,014 / 27,507 \\
\goodreads & book reviews & 34,010 / 17,005 \\
\amazon & reviews on amazon.com & 17,476 / 8,738\\
\gmane & newsgroups & 16,396 / 8,198 \\
\wikiDiscussions & Wikipedia editorial discussions & 13,724 / 6,862\\
\pubmed & medical journal articles & 692 / 346 \\
\bottomrule
\end{tabular}
\caption{Training data sources along with the number of document and authors selected from each source.}
\label{table:eng_8g}
\end{table*}

\section{Test data}
\label{appendix:test_data}

We benchmark our retrieve-and-rerank AA system on the HRS1 and HRS2 English HIATUS datasets (HRS stand for \textbf{H}IATUS \textbf{R}esearch \textbf{S}et).
These datasets are publicly available through the HIATUS website: https://www.iarpa.gov/research-programs/hiatus.  
Documents in HRS1 and HRS2 are drawn from five different genres.
The number of documents in each genre for both HRS1 and HRS2 are shown in Tables \ref{table:HRS1} and \ref{table:HRS2} respectively.
For each genre, we provide the number of documents in the foreground and the background collection.
Also, the websites from which the background documents are scraped are also shown.
Distribution of the number of documents in the foreground and background collections, along with unique number of authors is shown in Table \ref{table:foreground_background}.

The query documents in HRS1 and HRS2 are always drawn from the foreground collection.
To construct the query pool, we first sample a fraction of foreground authors.
For each selected author, one of the five genres is randomly chosen, and all of that author’s documents in the chosen genre are included as queries.
The same author’s documents in the remaining genres are placed in the candidate pool, ensuring evaluation under a cross-genre setting.
In addition, all background documents are added to the candidate collection.

Because the number of foreground authors and documents is limited (see Table \ref{table:foreground_background}), the sampling process typically yields fewer than 100 query documents.
To reduce variability, we construct $s=4$ distinct query–candidate splits, each generated with a different random seed (seed 0, 1001, 2001, 3001) when sampling query authors and genres.
System performance is then averaged across these splits, and the results are reported in Tables \ref{tab:retrieval} and \ref{tab:reranker}.
The exact number of queries and candidate documents in each split is provided in Table \ref{tab:seed_query_candidates}.

\begin{table*}[!th]
\centering
\begin{tabular}{@{}llcc@{}}
\toprule
Genre description &  Sources & \# Foreground docs & \# Background docs \\ \midrule
Board game reviews & boardmangeek.com & 212 & 2,432 \\
Citizens journalism & globalvoices.org & 140 & 5,453 \\
Instructions & instructables.com & 133 & 8,534 \\
Literature forums & stackexchange.com & 199 & 9,641 \\
STEM forums & stackexchange.com & 171 & 11,024 \\
\bottomrule
\end{tabular}
\caption{
Distribution of documents across genres in \textsc{hiatus}'s HRS1 test set.
}
\label{table:HRS1}
\end{table*}

\begin{table*}[!th]
\centering
\begin{tabular}{@{}llcc@{}}
\toprule
Genre description &  Sources & \# Foreground docs & \# Background docs \\ \midrule
Posts on pets & reddit.com & 112 & 1,298 \\
Personal tales and experiences & reddit.com & 138 & 1,542 \\
Product reviews & - & 121 & 0 \\
Obituary & multiple$\dagger$ & 132 & 3,625 \\
Movie reviews & letterboxd.com & 128 & 4,377 \\
\bottomrule
\end{tabular}
\caption{
Distribution of documents across genres in \textsc{hiatus}'s HRS2 test set.
$\dagger$The obituary posts are scraped from ballardsunderfuneral.com, 
dogwoodcremationcare.com, englandfuneralhome.com, pierfuneralhome.com, and snyderfuneralhome.com.
}
\label{table:HRS2}
\end{table*}

\begin{table*}[!th]
\centering
\begin{tabular}{@{}llcc@{}}
\toprule
HRS &  \# Foreground docs/authors & \# Background docs/authors \\ \midrule
HRS1 & 855/114 & 37084/16313  \\
HRS2 & 631/37 & 10842/2581  \\
\bottomrule
\end{tabular}
\caption{
Distribution of documents and unique authors in forground and background collection of HRS1 and HRS2.
}
\label{table:foreground_background}
\end{table*}

\begin{table*}[!th]
\centering
\begin{tabular}{@{}lcccc@{}}
\toprule
Sampling seed & \multicolumn{2}{c}{HRS1} & \multicolumn{2}{c}{HRS2} \\ 
\cmidrule(lr){2-3} \cmidrule(lr){4-5}
     & Query docs & Candidate docs & Query docs & Candidate docs \\ \midrule
0    & 119 & 34370 & 72  & 7357  \\
1001 & 112 & 34157 & 74  & 7393  \\
2001 & 104 & 33150 & 76  & 7464  \\
3001 & 129 & 31139 & 75  & 7365  \\
\bottomrule
\end{tabular}
\caption{
Number of query and candidate documents for HRS1 and HRS2 across different sampling seeds used to construct the query/candidate splits.
}
\label{tab:seed_query_candidates}
\end{table*}

\section{Training configuration}
\label{appendix:lora}

\textbf{Retriever hyperparameters}: All \sadiri and \iraa models are optimized using the Adam optimizer \cite{kingma2017adammethodstochasticoptimization} with a learning rate of $1e-5$ and a gradient accumulation step of 1. 
We set $\tau = 0.01$.
To be consistent with \cite{fincke2024separatingstylesubstanceenhancing}, we train the \sadiri models for 4 epochs with per-gpu batch-size of 74 authors.
To compensate for a larger parameter footprint when training LLMs, the \iraa models are only trained for a single epoch with per-gpu batch-size of 16 authors.
All retrievers are trained on a single node with 4 RTX-A6000 gpus.
All parameters are updated in bfloat16 and we use gradient checkpointing \cite{chen2016trainingdeepnetssublinear} to save memory while training. 
It takes roughly 15 hours to train our largest model (Mistral-12B). 

\textbf{Reranker hyperparameters}: All rerankers are finetuned versions of Mistral-Nemo-Base-2407\footnote{https://huggingface.co/mistralai/Mistral-Nemo-Base-2407} and are trained using the Adam optimizer \cite{kingma2017adammethodstochasticoptimization} with a learning rate of $1e-4$.
We set $\tau = 1.0$.
Each per-gpu training batch contains 1 query document.
For each query document, in addition to the single positive document, we select a total of $m = 12$ negative documents.
We use gradient accumulation of 10 steps.
All rerankers are trained on a single node with 4 RTX-A6000 gpus.
All parameters are updated in bfloat16 and we use gradient checkpointing \cite{chen2016trainingdeepnetssublinear} to save memory while training.
It takes roughly 60 hours to train each of the rerankers.

All retriever and reranker experiments involving Qwen3 and Mistral update only the LoRA parameters during training.
We leverage PEFT's\footnote{Version 0.17.1 of \url{https://github.com/huggingface/peft} installed via pip.} implementation of LoRA.
Table \ref{table:lora} shows the LoRA configuration that we used for all our experiments.

In addition to the LoRA parameters, we also train the projection matrix, $W$, in Eq. \ref{eq:retriever_score} and \ref{eq.reranker_score}.

\begin{table*}[!th]
\centering
\begin{tabular}{@{}ll}
\toprule
Property &  Value                                \\ 
\midrule
$r$             & 16                             \\
lora\_alpha     & $r\times2$                     \\
target\_modules & \{q,v,k,o,gate,up,down\}\_proj \\
lora\_dropout   & 0.05                           \\
bias            & none                           \\
task\_type      & FEATURE\_EXTRACTION            \\
\bottomrule
\end{tabular}
\caption{
PEFT's LoRA configuration which was used for all retriever and reranker experiments involving Qwen3 and Mistral.
}
\label{table:lora}
\end{table*}

\textbf{\sadiri's hard negative batching approach:}
\sadiri proposed to train the retriever by constructing training batches of similar authors.
This ensures that the in-batch negatives used in the contrastive loss are hard, providing better statistical pressure for the retriever to weed out the negatives in search of the positive document. 
The exact algorithm for this approach is shown in Algorithm \ref{alg:batch-construction}.
Consistent with \sadiri, we set the scaling factor $s = 3.5$ across all our experiments.
At the start of each epoch, the document embeddings from the current state of the retriever are used to cluster the entire training collection. 
Subsequently, the embeddings are clustered using K-means (via Faiss \cite{johnson2019billion}). 
Thus, this approach is memory intensive, requiring the embeddings of the entire training data to be loaded into GPU memory.
Since LLMs result in much larger document embeddings than RoBERTa-large, it becomes necessary to handle this memory requirement efficiently.

\textit{Efficient implementation using random projection}:
The clustering depends on cosine similarity of document embeddings, $\cos(v(d_1), v(d_2)) = v(d_1)^T v(d_2) / \|v(d_1)\| \|v(d_2)\|$.
We use this fact to reduce memory footprint by applying random matrix projection to each document embedding. 
This is motivated by the fact that \cite{Johnson1984ExtensionsOL} asserts that random projections preserve inner products.
Intuitively, let $r(d) = Rv(d_1)$ where $r \in R^K$ is the reduced dimensional vector that we get by projecting $v \in R^D$ using the random projection matrix $R \in R^{K \times D}$, $K < D$. 
Then
\begin{align*}
     r(d_1)^Tr(d_2) &= (Rv(d_1))^T(Rv(d_1)) \\
                    &= v(d_1)^TR^TRv(d_1).
\end{align*}
If each element in $R$ is drawn in an i.i.d from a univariate zero-mean distribution, then $R^T R \approx K \times \mathds{1}$.
With this assumption, we get $r(d_1)^Tr(d_2) \approx K v(d_1)^Tv(d_1)$, that is, the dot product of the projected vectors is approximately the same as the dot product of the original vectors (up to a scaling constant, $K$).
Since the scaling factor gets absorbed in normalization when calculating the cosine distance, we get $\cos(r(d_1), r(d_2)) \approx \cos(v(d_1), v(d_2))$.
For our experiments, we set $K = D // 3$.

\begin{algorithm}[t]
\caption{\sadiri's batching approach at the start of each epoch}
\label{alg:batch-construction}
\begin{algorithmic}[1]
\Require Training set of $N$ authors, each with $2$ documents: $\mathcal{D} = \{d_{i1}, d_{i2}\}_{i=1}^N$
\Require Embedding of all $2N$ documents
\Comment{Model output from previous epoch (For epoch 0, take output from the untrained model.)}
\Require Batch size $b$ authors, clustering factor $s$
\Ensure Similar author in the same batch
\State Cluster all $2N$ documents into $k = b \cdot s$ clusters
\For{each author $a_i$} 
    \Comment{Assign each author to a unique cluster}
    \If{documents $d_{i1}, d_{i2}$ are in different clusters}
        \State Keep the document in the larger cluster
    \EndIf
\EndFor
\For{clusters $C_j$ with $>b$ authors} 
    \Comment{Split/merge to respect batch size}
    \State Redistribute authors to smaller clusters
\EndFor
\State Randomly sample $s$ clusters to form each batch
\end{algorithmic}
\end{algorithm}

\end{document}